%% file: main.tex
\documentclass[10pt,twocolumn,letterpaper]{article}

\usepackage[pagenumbers]{cvpr} 


\usepackage{graphicx}
\usepackage{amsmath}
\usepackage{amssymb}
\usepackage{booktabs}
\usepackage{multirow}
\usepackage{array}
\newcolumntype{P}[1]{>{\centering\arraybackslash}p{#1}}

\AtBeginDocument{
\addtolength{\abovedisplayskip}{-2ex}
\addtolength{\abovedisplayshortskip}{-2ex}
\addtolength{\belowdisplayskip}{-2ex}
\addtolength{\belowdisplayshortskip}{-2ex}
}

\definecolor{cvprblue}{rgb}{0.21,0.49,0.74}
\usepackage[pagebackref,breaklinks,colorlinks,allcolors=cvprblue]{hyperref}

\def\paperID{0002} 
\def\confName{CVPR PVUW}
\def\confYear{2025}

\begin{document}

\title{\includegraphics[scale=0.05, trim=0 5cm -1cm 0]{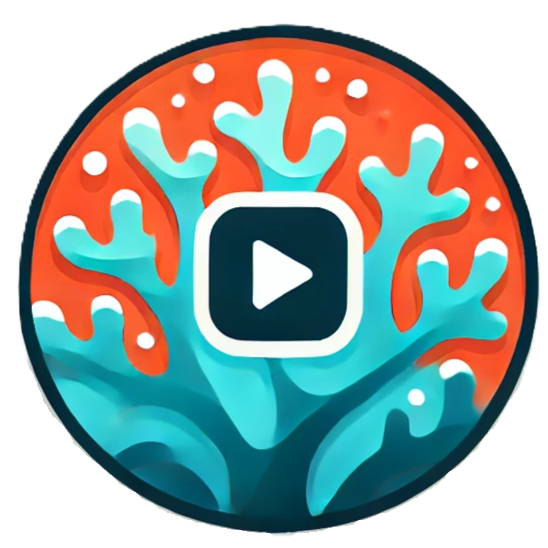}REEF: \underline{Re}levance-Aware and \underline{Ef}ficient LLM Adapter \\for Video Understanding}

\author{
    \hspace{-0.3cm} Sakib Reza$^{1}$\thanks{Work done during an internship at Futurewei Technologies Inc.} \hspace{0.3cm} Xiyun Song$^{2}$ \hspace{0.2cm} Heather Yu$^{2}$ \hspace{0.2cm} Zongfang Lin$^{2}$ \hspace{0.2cm} Mohsen Moghaddam$^{3}$ \hspace{0.2cm} Octavia Camps$^{1}$ \\
    \hspace{-0.4cm}$^1$Northeastern University \hspace{0.6cm} $^2$Futurewei Technologies Inc \hspace{0.6cm} $^3$Georgia Institute of Technology \\
    \hspace{-0.4cm}{\tt\footnotesize \{reza.s, o.camps\}@northeastern.edu,} {\tt\footnotesize \{xsong, hyu, zlin1\}@futurewei.com,} {\tt\footnotesize mohsen.moghaddam@gatech.edu} \\
}

\maketitle
\begin{abstract}
   Integrating vision models into large language models (LLMs) has sparked significant interest in creating vision-language foundation models, especially for video understanding. Recent methods often utilize memory banks to handle untrimmed videos for video-level understanding. However, they typically compress visual memory using similarity-based greedy approaches, which can overlook the contextual importance of individual tokens. To address this, we introduce an efficient LLM adapter designed for video-level understanding of untrimmed videos that prioritizes the contextual relevance of spatio-temporal tokens. Our framework leverages scorer networks to selectively compress the visual memory bank and filter spatial tokens based on relevance, using a differentiable Top-K operator for end-to-end training. Across three key video-level understanding tasks— untrimmed video classification, video question answering, and video captioning—our method achieves competitive or superior results on four large-scale datasets while reducing computational overhead by up to 34\%. The code will be available soon on GitHub.
\end{abstract}

\section{Introduction}
\label{sec:intro}
In recent years, Large Language Models (LLMs), such as GPT \cite{radford2018improving, radford2019language, NEURIPS2020_1457c0d6, gpt4} and LLaMA \cite{touvron2023llama, touvron2023llama2}, have demonstrated remarkable capabilities in text processing and generation. However, these models are inherently text-centric, limiting their ability to capture the multimodal nature of human interactions. To address this gap, recent research has extended LLMs to multimodal applications by integrating visual encoders, enabling image and video tasks such as visual captioning, visual question answering \cite{li2023blip2, liu2024visual, dai2023instructblip, zhu2024minigpt, ye2024mplug, zhang2023video, li2023videochat}, as well as visual classification, detection, and segmentation \cite{chen2024videollm, wang2024visionllm, chen2024minigptv, wang2022omnivl, qu2024llms, ren2024timechat}. Nevertheless, incorporating selective temporal information from untrimmed videos remains a major challenge in this area, limiting the full potential of multimodal systems.


\begin{figure}[!t]
	\centerline{\includegraphics[width=1\linewidth]{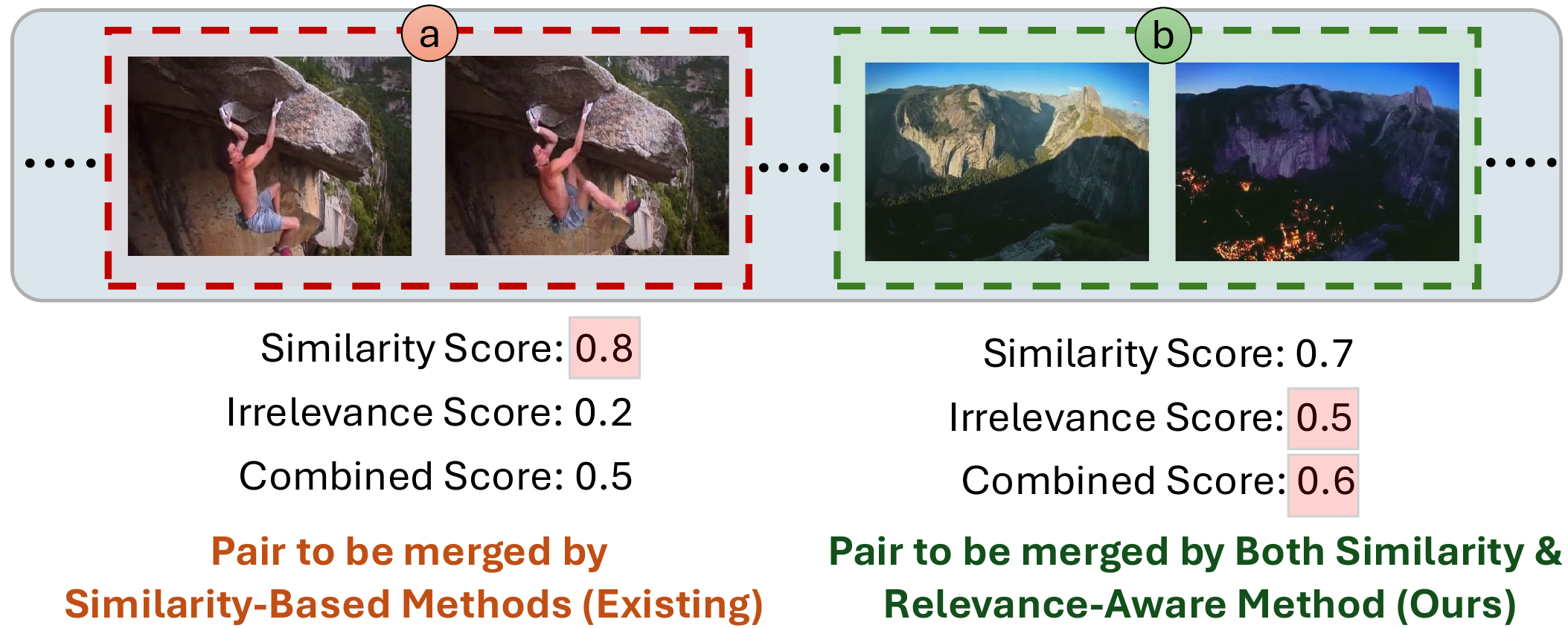}}
    \caption{Illustrative example of relevance-aware memory bank compression. In this example, the goal is to compress the memory bank by merging a pair of frames. While existing similarity-based memory compression methods \cite{he2024ma, song2024moviechat} would merge pair (a) due to its slightly higher similarity, this approach risks losing critical action information and retaining irrelevant context from (b). In contrast, our method makes more relevance-aware merging decisions, preserving the distinctiveness and contextual significance of the memory bank. Note that, in practice, memory banks consist of human non-interpretable feature tokens, not frames.} \label{fig:teaser}
    \vspace{-4mm}
\end{figure}

Previous large multimodal models \cite{li2023blip2} typically process video input by concatenating frame-wise query embeddings along the temporal axis before feeding them into LLMs. However, this method is constrained by the limited context length of LLMs and high GPU memory consumption, restricting the number of frames that can be efficiently processed. For instance, LLaMA has a context window of 2,048 tokens, while models like LLaVA \cite{liu2024visual} and BLIP-2 \cite{li2023blip2} can only accommodate 256 and 32 tokens per image, respectively, making direct application to untrimmed videos impractical. A straightforward alternative, such as temporal average pooling (as used in VideoChatGPT \cite{maaz2023videochatgpt}), often degrades performance by indiscriminately merging background, redundant, and contextually irrelevant frames with important ones. Video-LLaMA \cite{zhang2023video} partially mitigates this by employing a video querying transformer (Q-Former) to generate video-level representations, but this increases model complexity, making it unsuitable for real-time applications. To reduce computational overhead, recent methods such as MA-LMM \cite{he2024ma} and MovieChat \cite{song2024moviechat} process video frames sequentially, storing extracted features in a compressible memory bank while maintaining the architecture of existing large multimodal models \cite{li2023blip2, zhang2023video}. While this alleviates GPU memory constraints and LLM context length limitations, it remains suboptimal for preserving critical video information. Specifically, these memory banks rely on greedy compression strategies that prioritize adjacent similarity, often discarding broader contextual cues essential for video-level understanding of untrimmed videos.

To overcome this shortfall, we introduce \textbf{REEF}, a \textbf{RE}levance-aware and \textbf{EF}ficient LLM Adapter, which advances beyond simple memory bank compression by evaluating token relevance across the entire video. This ensures that essential information is retained without redundancy, addressing the contextual blind spots of earlier methods. By incorporating spatial token filtering, REEF not only improves the fidelity of untrimmed video representations but also significantly enhances computational efficiency. Thus, our approach builds on the strengths of existing multimodal models while addressing their key limitations in untrimmed video analysis. The main contributions of our work are: 

\begin{itemize}
    \item We present REEF, a novel LLM adapter to efficiently manage and process untrimmed video data for video-level understanding, which incorporates relevance-aware temporal compression and selective spatial filtering techniques.
    \item Hybrid Temporal Compression: REEF enhances video memory bank compression by considering both the relevance of frames within the video context and their visual similarity, leading to more meaningful data retention compared to existing methods (Figure \ref{fig:teaser}).
    \item Selective Spatial Filtering: By applying a spatial token selection technique, REEF effectively filters out redundant information, significantly improving the LLM adapter's efficiency in handling untrimmed video data.
    \item We conduct rigorous evaluations of our method across three video-level understanding tasks— untrimmed video classification, question answering, and captioning—using four large-scale datasets: Breakfast, ActivityNet, MSVD, and YouCook2. Our approach not only meets or exceeds state-of-the-art performance, but also achieves up to 34\% greater computational efficiency in terms of GFLOPs.
\end{itemize}

\section{Related Work}

\paragraph{Multi-Modal Large Language Models} 

Large Language Models (LLMs) \cite{NEURIPS2020_1457c0d6, chiang2023vicuna, gpt4, taori2023stanford, touvron2023llama, touvron2023llama2} have recently achieved remarkable success in natural language processing, spurring the development of Multimodal LLMs (MLLMs) \cite{alayrac2022flamingo, gong2023multimodal, li2023otter, li2023blip2, ye2024mplug, zhu2024minigpt}, which incorporate multiple data types. For example, Flamingo \cite{alayrac2022flamingo} bridges vision and language models, setting new standards in few-shot learning tasks. MiniGPT-4 \cite{zhu2024minigpt} aligns a visual encoder with a frozen LLM, while VideoChat \cite{li2023videochat} performs well in tasks requiring spatiotemporal reasoning. Vtimellm \cite{huang2024vtimellm} extends LLMs to interpret video content, and VideoLLM-Online \cite{chen2024videollm} introduces the LIVE framework for real-time video dialogue. Recent benchmarks such as MMT-Bench \cite{ying2024mmt} and ConvBench \cite{liu2024convbench} provide broader evaluations for these vision-language models. Video-LLaVA \cite{lin-etal-2024-video} first aligns visual features and projects them into the LLM. However, comprehension of untrimmed videos remains challenging due to high computational and memory demands. Koala \cite{tan2024koala} addresses this using learnable spatiotemporal queries for extended video analysis, while TimeChat \cite{ren2024timechat} employs time-stamped frames and an adaptive sliding window Q-Former for efficient compression. Models like MA-LMM \cite{he2024ma} and MovieChat \cite{song2024moviechat} incorporate memory mechanisms to improve understanding of long videos, though their compression approaches often lack sensitivity to contextual relevance.

\vspace{-0.5cm}
\paragraph{Untrimmed Video-Level Understanding}
Understanding the general context of untrimmed video poses significant challenges due to the presence of extensive background, noisy, and irrelevant frames, as well as spatial redundancies. Addressing this requires eliminating extraneous tokens while retaining essential information for effective representation learning. Prior to the emergence of LLM-based approaches, several methods tackled this problem by utilizing pre-extracted features instead of end-to-end joint backbone training \cite{papalampidi2024simple, donahue2015long, yue2015beyond, girdhar2017actionvlad, hussein2019timeception, wu2021towards}. Others adopted sparse sampling strategies, selecting key frames or employing cross-attention with learnable tokens to improve efficiency \cite{korbar2019scsampler, wu2019adaframe, reza2022history}. More recent advances, such as Vis4mer \cite{islam2022long} and S5 \cite{wang2023selective}, leverage the S4 \cite{guefficiently} transformer decoder for efficient temporal dependency modeling. LLM-based approaches for untrimmed video understanding can be categorized into retrieval-based and memory-based methods. Retrieval-based methods segment videos into smaller clips and extract relevant portions for LLM input \cite{fan2025videoagent, wang2024videoagent}. Memory-based models, inspired by prior memory-augmented techniques \cite{chen2020memory, lee2018memory, lee2021video, wu2022memvit}, integrate memory banks into large multimodal systems, as seen in MA-LMM \cite{he2024ma} and MovieChat \cite{song2024moviechat}. However, existing memory-based techniques rely heavily on similarity-based memory compression, which often fails to prioritize the most salient content over untrimmed sequences. This paper addresses these limitations by refining memory-based approaches to enhance focus on critical information in video-level understanding of untrimmed videos.

\vspace{-0.5cm}
\paragraph{Token Merging for Memory Compression}
Building on the token merging strategy introduced by \cite{bolya2023token}, several methods have adapted and extended this approach to address video data redundancy in video-language models. While these methods retain the core concept of merging similar tokens, they implement it in distinct ways. TESTA \cite{ren2023testa} employs a cascaded module for both spatial and temporal aggregation, progressively reducing video length and the number of tokens per frame. In contrast, Chat-UniVi \cite{jin2024chat} integrates modules that merge tokens across both dimensions in parallel before the large language model (LLM) processes the data. MovieChat \cite{song2024moviechat} applies a selective strategy, merging similar adjacent frames to decrease the total number of video frames. Likewise, MA-LMM \cite{he2024ma} performs token merging primarily along the temporal dimension but with finer granularity at the spatial level, independently compressing visual and query tokens across various spatial areas to enhance performance. However, merging the most visually similar token pairs alone may be suboptimal, as it overlooks their contextual relevance within the video. To address this, we propose a novel approach that combines a similarity-based greedy algorithm with a relevance-based method, improving token merging for memory bank compression.

\section{Method}

\begin{figure*}[htbp]
	\centerline{\includegraphics[width=1\linewidth]{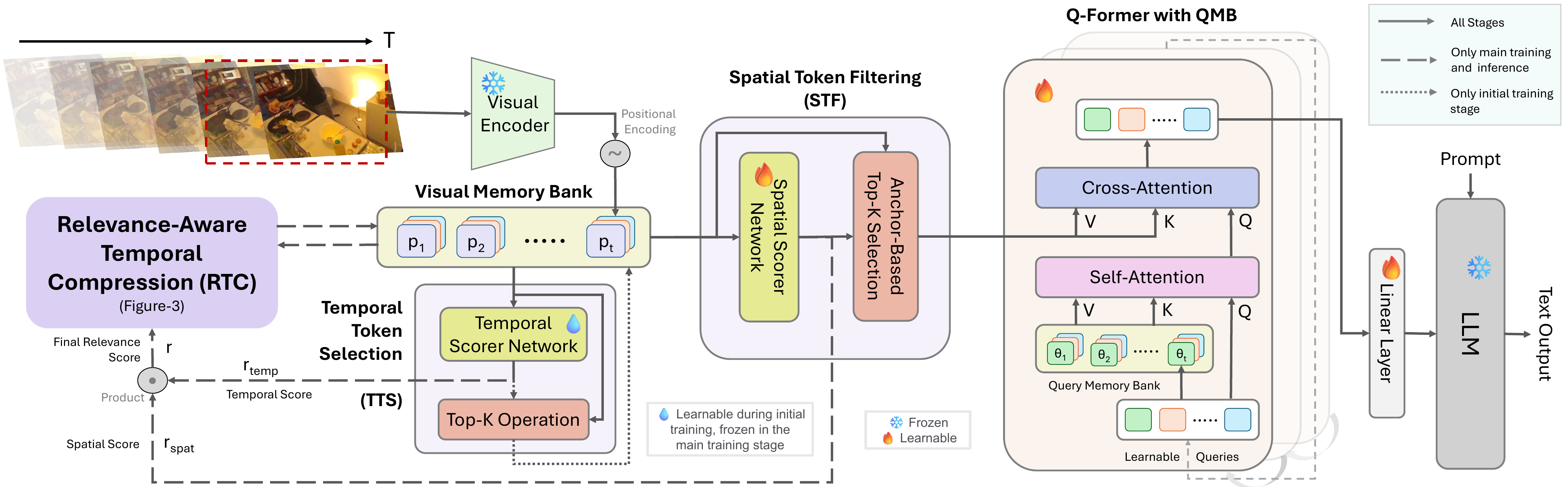}}
    \caption{The proposed LLM-based REEF framework for video understanding. Encoded frame visual features are stored in the visual memory bank over time, with the STF module removing redundant and irrelevant spatial information. These filtered features are then sent to the Q-Former for temporal modeling and alignment with the text domain. The RTC module collects token relevance scores from both the TTS and STF modules, compressing the memory bank by preserving only the most relevant and distinctive information based on adjacent similarity and relevance, ensuring the memory size remains fixed.}
    \label{fig:model_overview}
    \vspace{-4mm}
\end{figure*}

We introduce REEF, a relevance-aware and selective LLM adapter designed for efficient untrimmed holistic video understanding. Unlike traditional approaches \cite{he2024ma, song2024moviechat} that rely on similarity-based greedy methods for processing visual memory banks, our method employs relevance-based visual memory compression, enabling more meaningful retention of essential information. By implementing spatial token filtering, REEF also eliminates redundant spatial information, significantly enhancing its processing efficiency. Figure \ref{fig:model_overview} provides an overview of our LLM-based REEF framework, which is described in detail across several key sections: Sec \ref{sec:feat} covers visual feature extraction using a pre-trained, frozen visual encoder and Sec \ref{sec:temp_modeling} delves into temporal modeling and visual-text alignment with a querying transformer (Q-Former).  Sec \ref{sec:rtc} details one of our core contributions: the relevance-aware temporal compression module. Sec \ref{sec:spat} describes the spatial token filtering method while Sec \ref{sec:training} outlines our two-stage framework training strategy, including initial training to set up the temporal relevance scorer network using a differentiable Top-K selection method, and the main training phase.

\subsection{Initial Feature Extraction} 
\label{sec:feat}
To begin with, we extract spatial features from the input video frames for further processing and downstream tasks. Given a sequence of \( T \) video frames, we pass them through a pre-trained visual encoder, such as a Vision Transformer (ViT), to obtain visual features represented as \( \mathbf{E}_t = [\mathbf{e}_1, \mathbf{e}_2, \dots, \mathbf{e}_t] \), where \( \mathbf{e}_t \in \mathbb{R}^{N \times D} \) are the features of the \( t \)-th frame. Here, \( N \) denotes the number of patches per frame, and \( D \) is the channel dimension.

To account for the temporal order of the frames, we combine the visual features of each frame with their  positional embedding \( \text{PE}(t) \): 
$\mathbf{p}_t = \mathbf{e}_t + \text{PE}(t)$, 
$\mathbf{p}_t \in \mathbb{R}^{N \times D}$.
This process effectively integrates spatial and temporal order information, preparing the sequence for tasks that require an understanding of video dynamics.

\subsection{Temporal Modeling}
\label{sec:temp_modeling}
To align the visual and text embeddings, we utilize the same architecture as the Querying Transformer (Q-Former) \cite{li2023blip2} connected to the visual and query memory banks from MA-LMM \cite{he2024ma}. The Q-Former leverages learned queries \( \boldsymbol{\theta} \in \mathbb{R}^{L \times D} \) to capture temporal information in videos, where \( L \) represents the number of learned queries. Each Q-Former block comprises two attention submodules: (1) a cross-attention layer that interacts with the raw visual features extracted from the pre-trained visual encoder, and (2) a self-attention layer that models interactions among the input queries. Unlike the original Q-Former in BLIP-2, which only attends to the current frame's embedding, this design incorporates a video-level memory bank, consisting of both the visual memory bank and the query memory bank. This memory bank accumulates past video information, enhancing the input to the cross- and self-attention layers for effective untrimmed video understanding.

\noindent \textbf{Visual Memory Bank.}
It retains raw visual features from each frame, extracted by a pre-trained visual encoder. At time step \( t \), the Visual Memory Bank stores a concatenated sequence of visual features \( \mathbf{P}_t = \text{Concat}[\mathbf{p}_1, \mathbf{p}_2, \dots, \mathbf{p}_t] \), where \( \mathbf{P}_t \in \mathbb{R}^{tN \times D} \). For an input query \( \boldsymbol{\theta}_t \), the Visual Memory Bank functions as both the key and value in the attention mechanism: \( \mathbf{Q} = \boldsymbol{\theta}_tW_Q, \quad \mathbf{K} = \mathbf{P}_t\mathbf{W}_K, \quad \mathbf{V} = \mathbf{P}_t\mathbf{W}_V. \) Cross-attention is then computed as: \( \mathbf{O} = \text{Attn}(\mathbf{Q}, \mathbf{K}, \mathbf{V}) = \text{Softmax} \left(\frac{\mathbf{QK}^T}{\sqrt{C}}\right) \mathbf{V}. \) This mechanism enables the model to explicitly attend to historical visual information stored in the Visual Memory Bank, leveraging long-term contextual dependencies. Notably, all cross-attention layers within the Q-Former share access to the same Visual Memory Bank, ensuring consistent access to visual features across all Q-Former blocks.

\noindent \textbf{Query Memory Bank.} 
In addition to the Visual Memory Bank, following \cite{he2024ma}, the Query Memory Bank accumulates learned queries over time, refining the model’s evolving understanding of each frame. Unlike the static Visual Memory Bank, it captures dynamic temporal dependencies across frames. Further details on the Query Memory Bank can be found in the supplementary material section \ref{app:qmb}.

\subsection{Relevance-Aware Temporal Compression}
\label{sec:rtc}
In video processing, storing past frames in memory causes GPU usage and computation to grow linearly, limiting scalability for untrimmed videos. To address this, methods like MeMViT \cite{wu2022memvit} use a first-in-first-out (FIFO) queue to cap memory size by discarding older features, though this loses long-term context. MeMViT also uses learnable pooling to compress spatio-temporal features, but this adds extra parameters and complexity.

Building on the success of token merging and pruning techniques \cite{bolya2023token, yin2022vit, meng2022adavit, rao2021dynamicvit}, Bo He et al. \cite{he2024ma} proposed a Memory Bank Compression (MBC) technique  that leverages temporal redundancies in video data by aggregating and compressing similar adjacent features. This method effectively retains earlier historical information while reducing redundant data within the memory bank. However, MBC's focus on redundancy reduction may not sufficiently filter out irrelevant or background information while preserving critical content. To address this, we propose a new approach that introduces relevancy score-based context-aware memory bank compression. Our method assigns importance scores to each token within the video context and uses these scores, along with the similarity between adjacent frames, to more effectively aggregate and compress video information, ensuring that contextually significant content is preserved while irrelevant data is filtered out.
\vspace{-0.6cm}
\paragraph{Relevance Scorer Network}
To evaluate the relevance of each spatio-temporal token, we use a lightweight neural network designed to generate importance scores. Given input tokens \( \mathbf{x} \in \mathbb{R}^{(L+1) \times N \times D} \), the objective of temporal selection is to choose \( K \) of the \( L+1 \) frames and discard the rest. First, we apply average pooling to \( \mathbf{x} \) along the spatial dimension to obtain a sequence of frame-based tokens \( \mathbf{x}_t \in \mathbb{R}^{(L+1) \times D} \), which are then processed by the scorer network. The goal of the scorer network is to produce input-conditioned importance scores for each token.

The scorer network uses a standard two fully-connected linear layer architecture. Initially, the input tokens are transformed into a local representation \( \mathbf{f}^l \) through a linear projection: \( \mathbf{f}^l = \text{Linear}(\mathbf{x}_t; \boldsymbol{\omega}_1) \in \mathbb{R}^{(L+1) \times D'} \), where \( \boldsymbol{\omega}_1 \) represents the network's weights, and \( D' = D/2 \) is used to reduce computational complexity. To incorporate contextual information across the entire sequence, the local representations \( \mathbf{f}^l \) are averaged to generate a global representation \( \mathbf{f}^g \). This global representation is then concatenated with each local representation along the channel dimension: \( \mathbf{f}_i = [\mathbf{f}_i^l, \mathbf{f}^g] \in \mathbb{R}^{2D'} \), where \( 1 \leq i \leq L+1 \). These concatenated features \( \mathbf{f}_i \) are passed through a second linear layer, which produces unnormalized relevance scores for temporal tokens: \(\mathbf{r}'_{temp} = \text{Linear}(\mathbf{f}; \omega_2) \in \mathbb{R}^{(L+1) \times 1} \), where \( \boldsymbol{\omega}_2 \) represents the weights of the second linear layer. Finally, min-max normalization is applied to these scores to obtain the final importance scores: \( \mathbf{r}_{temp} = \text{Norm}(\mathbf{r}'_{temp}) \), where \( \mathbf{r}_{temp} \in \mathbb{R}^{L+1} \) is the vector of normalized scores for all tokens. This scorer network introduces minimal computational overhead, which is significantly outweighed by the efficiency gains from pruning uninformative tokens.

\begin{figure}[!t]
	\centerline{\includegraphics[width=1\linewidth]{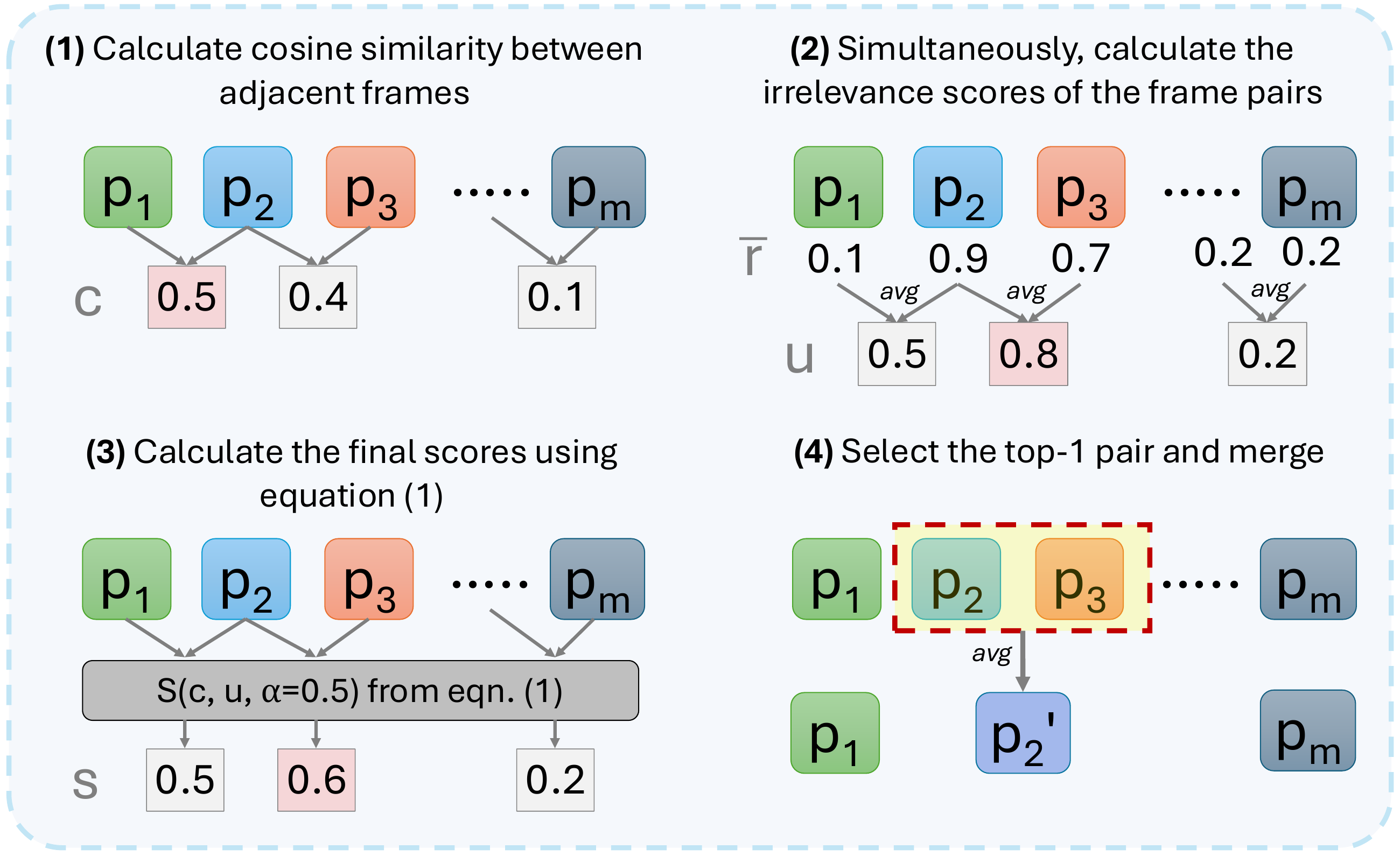}}
    \caption{Steps of the proposed Relevance-aware Temporal Compression (RTC). This approach preserves the most discriminative and relevant features while maintaining the temporal order.} \label{fig:rtc}
    \vspace{-4mm}
\end{figure}

\vspace{-0.5cm}
\paragraph{Final Score Calculation and Merging}
Figure \ref{fig:rtc} provides a high-level overview of our compression operation.
Similar to MBC \cite{he2024ma}, REEF applies the compression algorithm at each autoregressive step if the current length of the memory bank exceeds the predefined threshold \( L \). Formally, consider a visual memory bank containing a list of features \([\mathbf{p}_1, \mathbf{p}_2, \dots, \mathbf{p}_L]\), where \( \mathbf{p}_t \in \mathbb{R}^{N \times D} \). When a new frame feature \( \mathbf{f}_{L+1} \) arrives, we need to compress the memory bank by reducing its length by one. 

At each spatial location \( i \), we calculate the cosine similarity between temporally adjacent tokens: \( c^i_t = \cos(p^i_t, p^i_{t+1}) \) for \( t \in [1, L] \), \( i \in [1, N] \). Simultaneously, we compute a relevancy score for each temporal token \( \mathbf{r}_{\text{temp}} \) (described previously), and get the relevancy score for the spatial token \( \mathbf{r}_{\text{spat}} \) from the spatial token filtering module (more details in section \ref{sec:spat}). These are multiplied together to get the final spatio-temporal relevance probability: \( \mathbf{r} = \mathbf{r}_{\text{temp}} \cdot \mathbf{r}_{\text{spat}} \in \mathbb{R}^{(L+1) \times N} \), where \( \mathbf{r} \) represents the overall spatio-temporal importance or relevance probability.

We then complement it to get the irrelevance probability \( \Bar{\mathbf{r}} = 1 - \mathbf{r}\) for \( \Bar{\mathbf{r}}^i_t \in [0, 1] \), and calculate the average irrelevancy probability between temporally adjacent tokens: \( \mathbf{u}_t = \frac{\Bar{\mathbf{r}}_t + \Bar{\mathbf{r}}_{t+1}}{2} \).

Next, we combine these scores using a weighted average, controlled by a hyperparameter \( \alpha \in [0, 1] \), resulting in the final score for each token:
\begin{align}
s^i_t = \alpha \cdot c^i_t + (1 - \alpha) \cdot u^i_t.
\label{eqn:alpha}
\end{align}
We then select the highest-scoring token across time, which corresponds to the most temporally redundant and irrelevant features: \( k = \arg\max_t(s^i_t) \).

To compress the memory bank, we average the selected token pair's features across all spatial locations, reducing the memory bank length by one: \( \hat{p}^i_k = \frac{p^i_k + p^i_{k+1}}{2} \). 


\subsection{Spatial Token Filtering}
\label{sec:spat}
Spatial selection is applied independently to each frame, where the objective is to select \(K\) out of \(N\) tokens per frame. We begin by processing the tokens of all frames in the memory bank \(\mathbf{x} \in \mathbb{R}^{(L+1) \times N \times D}\) through a scoring network (similar to the scorer network in section \ref{sec:rtc}) to obtain relevance scores \(\mathbf{r}_{spat} \in \mathbb{R}^{(L+1) \times N}\). A naive approach to select the top \(K\) tokens is to directly apply the Top-K operator to the scores \(\mathbf{r}_{spat}\). However, this disrupts the spatial coherence of the tokens, which is particularly detrimental in video transformers. 

\begin{figure}[htbp]
	\centerline{\includegraphics[width=0.9\linewidth]{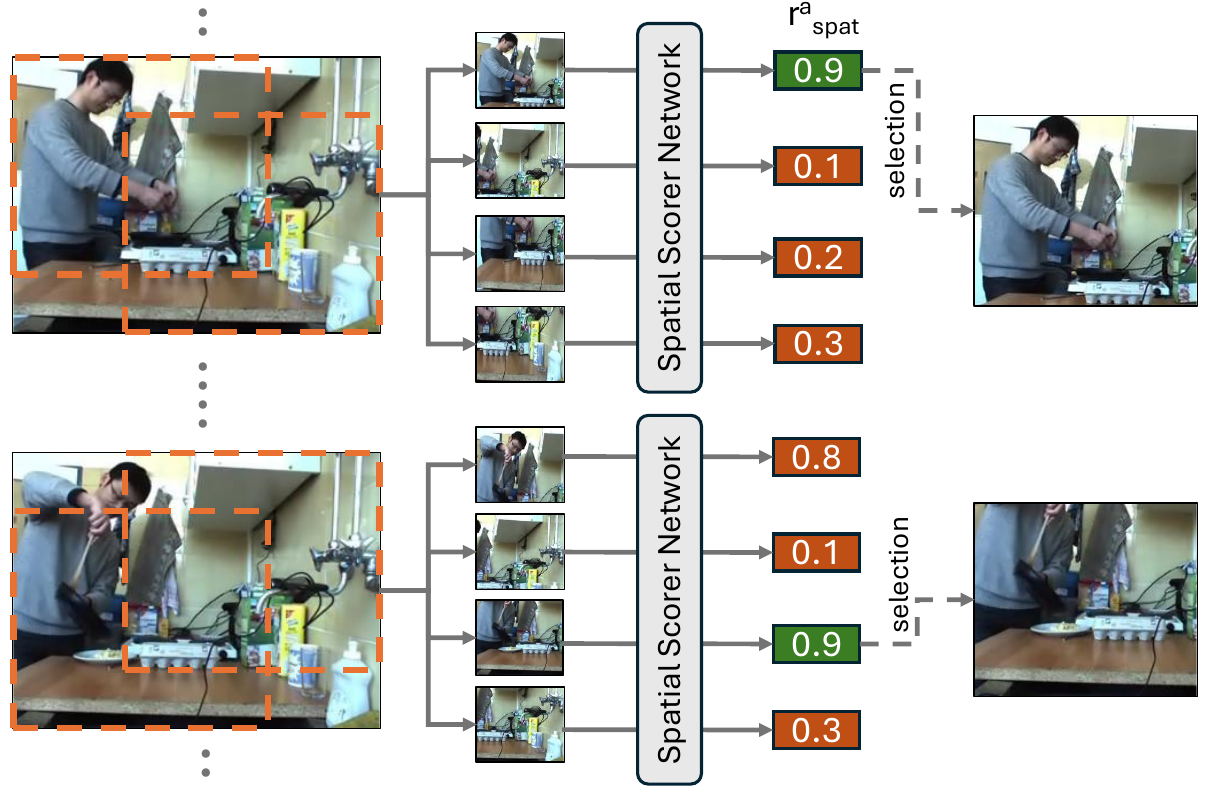}}
    \vspace{-2mm}
    \caption{Conceptual overview of our spatial token filtering method. This approach preserves the most discriminative and relevant regions based on predefined anchors. In practice, the filtering is applied to spatial tokens within the temporal memory bank, rather than directly on the video frames.} \label{fig:spat}
    \vspace{-4mm}
\end{figure}


We utilize an anchor-based approach for spatial selection similar to \cite{wang2022efficient}. After generating importance scores \(\mathbf{r_{{spat}_{t}}} \in \mathbb{R}^N\) for each frame at time step \(t\), we reshape the scores into a 2D map \(\mathbf{r}^s_{\text{spat}} \in \mathbb{R}^{\sqrt{N} \times \sqrt{N}}\). For simplicity, we omit the time step subscript \(t\) in the remainder of the formulation. The 2D map is then segmented into a grid of overlapping anchors \(\mathbf{r}^a_{\text{spat}} \in \mathbb{R}^{H \times K}\), where \(H = \left(\frac{\sqrt{N} - \sqrt{K}}{\gamma} + 1\right)^2\) denotes the number of anchors, given anchor size \(K\) and stride \(\gamma\). We aggregate the scores within each anchor using average pooling to obtain anchor-level scores \(\mathbf{r}^a_{\text{spat}} \in \mathbb{R}^H\). This reformulates the Top-K selection as a Top-1 selection task, where the anchor with the highest score is selected using a Top-K operator with \(K = 1\). Figure \ref{fig:spat} illustrates a simplified example of this method.


\subsection{Framework Training}
\label{sec:training}

REEF is trained in two stages: initial and main training. 
During the initial stage, we focus on training the temporal scorer network, which is critical for the subsequent relevance-aware temporal compression. This stage involves the Top-K temporal token selection (TTS) process, where the temporal token selection module selects the top \(K=L\) tokens from a set of \(L+1\) tokens within the memory bank. This selection process compresses the memory bank, after which the compressed tokens are passed through the Q-Former, with the remaining processing steps following the procedure outlined earlier. Notably, our empirical findings indicate that this initial stage requires significantly fewer epochs than the main training stage. 

In the main training phase, the pre-trained temporal scorer network is integrated into the relevance-aware temporal compression module. During this phase, the visual encoder, the temporal scorer network, and the large language model (LLM) are kept frozen, while the remaining components of the framework are trained. This allows the system to fine-tune the learnable components effectively, leveraging the pre-trained knowledge from the earlier stage.
\vspace{-0.5cm}
\paragraph{Text Decoding and Loss Function.}
REEF adopts a text decoding method similar to that described in \cite{he2024ma}. Specifically, we process video frames in an auto-regressive manner, where the Q-Former output at the final timestep encapsulates all previous information and is then fed into the LLM. This strategy allows us to reduce the number of input text tokens from \(M \times T\) to \(M\), effectively addressing the context length limitations of current LLMs and significantly reducing GPU memory requirements. 

During training, our model is supervised using a standard cross-entropy loss, with a labeled dataset comprising video-text pairs. The loss function is defined as:
\begin{align}
L = -\frac{1}{S} \sum_{i=1}^{S} \log P(\psi_i \mid \psi_{<i}, V)
\end{align}
where \(V\) represents the input video, and \(\psi_i\) is the \(i\)-th ground-truth text token. In this process, we update the parameters of the Q-Former while keeping the weights of both the visual encoder and the language model frozen, following the techniques outlined in the referenced works.

\vspace{-0.3cm}
\paragraph{Differentiable Top-K Selection.}
In our framework, the temporal and spatial scorer networks are trained using a Top-K token selection strategy. Specifically, given the relevance scores \( \mathbf{r'} \) generated by the scorer network, we select the \( K \) highest scores and extract the corresponding tokens. This process, denoted as a Top-K operator, returns the indices of the \( K \) largest entries: \( \mathbf{y} = \text{Top-K}(\mathbf{r'}) \in \mathbb{N}^K \). We assume that the indices are sorted to preserve the order of the input sequence. To implement token selection using matrix multiplication, we convert \( \mathbf{y} \) into a stack of \( K \) one-hot \( G \)-dimensional vectors \( \mathbf{Y} = [\mathbf{I}_{y_1}, \mathbf{I}_{y_2}, ..., \mathbf{I}_{y_K}] \in \{0,1\}^{G \times K} \), where \( G \) is the total number of initial tokens. As a result, tokens with top \( K \) scores can be extracted as \( \mathbf{x}' = \mathbf{Y}^\top \mathbf{x} \). Note that this operation is non-differentiable because both Top-K and one-hot operations are non-differentiable.

To train the parameters of the scorer network using an end-to-end approach without introducing any auxiliary losses, we use the perturbed maximum method \cite{berthet2020learning, wang2022efficient, cordonnier2021differentiable} to construct a differentiable Top-K operator. In particular, selecting Top-K tokens is equivalent to solving a linear program of the following form:
\begin{align}
\arg\max_{\mathbf{Y} \in \mathcal{C}} \langle \mathbf{Y}, \mathbf{r'} \mathbf{1}^\top \rangle,
\label{eqn:perturbed}
\end{align}
where \( \mathbf{s} \mathbf{1}^\top \) is the score vector replicated \( K \) times, and \( \mathcal{C} \) is the convex polytope constraint set defined as:
\vspace{0.15cm}
 \[ 
\mathcal{C} = \left\{ \mathbf{Y} \in \mathbb{R}^{G \times K} : \mathbf{Y}_{g,k} \geq 0, \mathbf{1}^\top \mathbf{Y} = 1, \mathbf{Y} \mathbf{1} \leq 1, \right.
\]
\vspace{0.02cm}
\[ 
\left. \sum_{i \in [G]} i \mathbf{Y}_{i,k} < \sum_{j \in [G]} j \mathbf{Y}_{j,k'}, \forall k < k' \right.\rule[-2ex]{0em}{6ex}\}. \tag{4}
\]

\vspace{0.15cm}


Following \cite{wang2022efficient}, we perform forward and backward operations to solve Equation \ref{eqn:perturbed}, with additional details in the Supplementary Material Section \ref{app:topk}.

\section{Experiments}

\subsection{Evaluation Tasks and Datasets}
To validate the effectiveness of the proposed REEF, we evaluate it on three video understanding tasks: untrimmed video classification, video question answering, and video captioning, comparing our approach against existing methods.

\noindent \textbf{Untrimmed Video Classification.}
We conduct experiments on two untrimmed video classification datasets: Breakfast \cite{kuehne2014language} and ActivityNet\cite{caba2015activitynet}. We use top-1 classification accuracy as the evaluation metric. The Breakfast dataset [56] comprises 1,712 videos focused on breakfast preparation, with an average video length of approximately 2.7 minutes. ActivityNet includes 14,950 untrimmed videos, with an average duration of around 2 minutes. Since the videos in these datasets are untrimmed, they include numerous background and irrelevant frames, making the problem more challenging and distinct from standard clipped video classification.

\noindent \textbf{Video Question Answering.}
We evaluate our model on two open-ended video question-answering datasets: MSVD-QA \cite{xu2017video} and ActivityNet-QA \cite{yu2019activitynet}. The ActivityNet-QA dataset shares the same videos as ActivityNet, but includes additional QA annotations. 

\noindent \textbf{Video Captioning.}
We report video captioning results using the METEOR \cite{banerjee2005meteor} and CIDEr \cite{vedantam2015cider} metrics on two widely-used datasets: MSVD \cite{chen2011collecting} and YouCook2 \cite{zhou2018towards}.

In addition to the standard evaluation metrics, we report the estimated GFLOPs of the LLM adapter for our proposed method and the baseline LLM-based approaches to compare computational efficiency. Further details on GFLOPs calculation can be found in the Supplementary Material Section \ref{app:gflops}.

\subsection{Implementation Details} 
We employ the pre-trained image encoder ViT-G/14 \cite{dosovitskiy2021an} from EVA-CLIP \cite{fang2023eva} for the visual encoder, with the option to substitute it with other CLIP-based video encoders. The pre-trained Q-Former weights from InstructBLIP \cite{dai2023instructblip} are used, and Vicuna-7B \cite{chiang2023vicuna} serves as the LLM. All experiments are run on four NVIDIA H100 GPUs. For hyperparameters, the score balance coefficient \(\alpha\) is set to 0.7 across all datasets and tasks, except for ActivityNet, where \(\alpha = 0.9\) is used for both video classification and question answering. The number of selected spatial tokens is 100 for all datasets and tasks, except for ActivityNet (144 for classification and 121 for QA) and YouCook2 (144 for video captioning). Additional training and evaluation details can be found in the Supplementary Material Section \ref{app:config}.

\subsection{Comparison with State-of-the-Art Methods}

We compare the performance of  REEF against  state-of-the-art (SOTA) methods in video-level understanding benchmarks for untrimmed videos. In most cases, our method either surpasses the SOTA across various metrics or achieves comparable results with significantly improved computational efficiency, demonstrating the effectiveness of our approach.

Table \ref{Tab:class} presents untrimmed video classification results on two large-scale datasets: Breakfast and ActivityNet. For the Breakfast dataset, REEF notably outperforms the baseline MA-LMM and other methods in terms of accuracy, while  offering approximately 34\% greater efficiency in terms of GFLOPs. On the ActivityNet dataset, REEF achieves a slight improvement in accuracy compared to SOTA, with a marked reduction in computational overhead.

Table \ref{Tab:qa} shows video question answering results on the MSVD-QA and ActivityNet-QA datasets. For MSVD-QA, REEF surpasses the SOTA in terms of accuracy and significantly improves computational efficiency. On ActivityNet-QA, REEF matches the accuracy of the baseline while reducing GFLOPs, confirming enhanced efficiency.

In Table \ref{Tab:cap}, we present video captioning results on the MSVD and YouCook2 datasets. For MSVD, our method significantly outperforms SOTA on the CIDEr metric while also being more computationally efficient, and achieves near-parity on METEOR. On YouCook2, our model achieves a slight improvement in METEOR and better efficiency in terms of GFLOPs, though it falls short on CIDEr compared to SOTA.

\begin{table}

\caption{Comparison of untrimmed video classification performance on the Breakfast and ActivityNet datasets, in terms of accuracy (\%) and computational efficiency (GFLOPs).}
\vspace{-0.2cm}
\label{Tab:class}
\centering
\resizebox{0.47\textwidth}{!}{
\begin{tabular}{l | c c | c c} 
 \toprule
  Model &  \multicolumn{2}{c|}{Breakfast} & \multicolumn{2}{c}{ActivityNet} \\ [0.5ex] 
   \scriptsize{*reconstructed results}& Acc.$\uparrow$ & GFLOPs$\downarrow$ & Acc.$\uparrow$ & GFLOPs$\downarrow$ \\
 \hline
    VideoGraph \cite{hussein2019videograph} & 69.5 & \multirow{6}{*}{N/A}  & - &  \multirow{6}{*}{N/A}
    \\
    Timeception \cite{hussein2019timeception} & 71.3 &  & 69.4 & 
    \\
    GHRM \cite{zhou2021graph} & 75.5 &  & - & 
    \\
    D-Sprv. \cite{lin2022learning} & 89.9 &  & - & 
    \\
    ViS4mer \cite{islam2022long} & 88.2 &  & 87.1 & 
    \\
    S5 \cite{wang2023selective} & 90.7 &  & - & 
    \\
    \bottomrule
    
    Video-LLaMa \cite{zhang2023video} & 90.7 & \underline{151.4} & 90.2 & \underline{151.4}
    \\
    MA-LMM\small{*} \cite{he2024ma} & \underline{91.6} & 154.7 & \underline{91.5} & 154.7
    \\
    \textbf{REEF (Ours)} & \textbf{93.8} & \textbf{102.3} & \textbf{91.6} & \textbf{126.8} 
    \\
 
\bottomrule
\end{tabular}
}
\end{table}

\begin{table}

\caption{Comparison of video question answering performance on the MSVD and ActivityNet datasets, in terms of top-1 accuracy (\%) and computational efficiency (GFLOPs).}
\vspace{-0.2cm}
\label{Tab:qa}
\centering
\resizebox{0.47\textwidth}{!}{
\begin{tabular}{l | c c | c c} 
 \toprule
  Model &  \multicolumn{2}{c|}{MSVD} & \multicolumn{2}{c}{ActivityNet} \\ [0.5ex] 
   \scriptsize{*reconstructed results}& Acc.$\uparrow$ & GFLOPs$\downarrow$ & Acc.$\uparrow$ & GFLOPs$\downarrow$ \\
 \hline
    JustAsk \cite{yang2021just} & 47.5 & \multirow{8}{*}{N/A} & 38.9 & \multirow{8}{*}{N/A} \\
    FrozenBiLM \cite{yang2022zero} & 54.8 & & 43.2 & \\
    SINGULARITY \cite{lei2023revealing} & - & & 44.1 &  \\
    VIOLETv2 \cite{fu2023empirical} & 54.7 & & - &  \\
    GIT \cite{wanggit} & 56.8 & & - &  \\
    mPLUG-2 \cite{xu2023mplug} & 58.1 & & - &  \\
    UMT-L \cite{li2023unmasked} & 55.2 & & 47.9 &  \\
    VideoCoCa \cite{yan2022videococa} & 56.9 & & \textbf{56.1} & \\
    \bottomrule
    
    Video-LLaMa \cite{zhang2023video} & 58.3 & \underline{83.1} & 45.5 & \underline{83.1}
    \\
    MA-LMM\small{*} \cite{he2024ma} & \underline{59.4} & 86.4 & \underline{48.7} & 86.4
    \\
    \textbf{REEF (Ours)} & \textbf{59.9} & \textbf{59.0} & \underline{48.7} & \textbf{65.1}
    \\
 
\bottomrule
\end{tabular}
}
\vspace{-0.5cm}
\end{table}

\begin{table}

\caption{Comparison of video captioning performance on the MSVD and YouCook2 datasets, in terms of METEOR (M), CIDEr (C), and computational efficiency in GFLOPs (GF).}
\vspace{-0.2cm}

\label{Tab:cap}
\centering
\resizebox{0.47\textwidth}{!}{
\begin{tabular}{l | c c c | c c c } 
 \toprule
  Model & \multicolumn{3}{c|}{MSVD} & \multicolumn{3}{c}{YouCook2}\\ [0.5ex] 
    & M $\uparrow$ & C $\uparrow$ & GF $\downarrow$ & M $\uparrow$ & C $\uparrow$ & GF $\downarrow$ \\ 
 \hline
    UniVL \cite{luo2020univl} & 29.3 & 52.8 &  \multirow{5}{*}{N/A} &  & 127.0 &  \multirow{5}{*}{N/A} \\
    SwinBERT \cite{lin2022swinbert} & 41.3 & 120.6 & & 15.6 & 109.0 &  \\
    GIT \cite{wanggit} & \textbf{51.1} & \underline{180.2} & & 17.3 & \underline{129.8} &  \\
    mPLUG-2 \cite{xu2023mplug} & 48.4 & 165.8 &  & - & - &  \\
    VideoCoca \cite{yan2022videococa} & - & - &  & - & 128.0 & \\
    \bottomrule
    
    Video-LLaMA \cite{zhang2023video} & 49.8 & 175.3 & \underline{287.9} & 16.5 & 123.7 & \underline{287.9} 
    \\
    MA-LMM \cite{he2024ma} & \underline{51.0} & 179.1 & 291.2 &  \underline{17.6} & 
    \textbf{131.2} & 291.2
    \\
    \textbf{REEF (Ours)} & 50.5 & \textbf{191.6} & \textbf{189.2} & \textbf{17.7} & 126.8 & \textbf{237.1}
    \\

\bottomrule
\end{tabular}
}
\vspace{-0.4cm}
\end{table}

\begin{figure*}[htbp]
	\centerline{\includegraphics[width=1\linewidth]{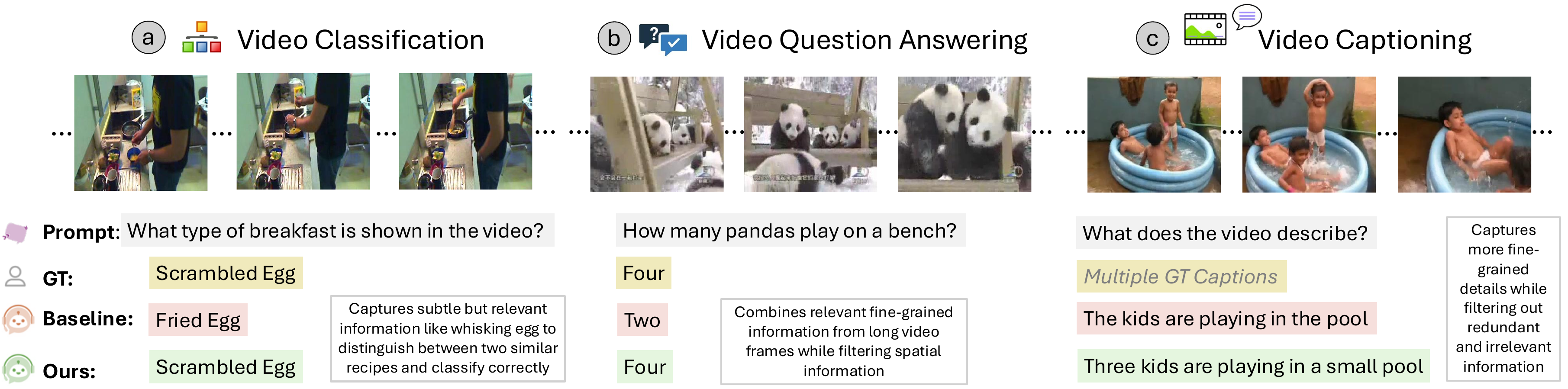}}
    \caption{Qualitative evaluation of our REEF model across three video 
 understanding tasks, compared to the baseline MA-LMM.} \label{fig:qual}
    \vspace{-4mm}
\end{figure*}

Additionally, we conducted a qualitative analysis to further evaluate the effectiveness of our model. While, in most cases, our model performs comparably to the baseline MA-LMM model, there are instances where it outperforms. For instance, in Figure \ref{fig:qual}(a) for video classification, the baseline struggled to distinguish between the subtle differences in actions like making scrambled eggs versus frying eggs, whereas our model successfully differentiated between them. In Figure \ref{fig:qual}(b), during video question answering, our model accurately counted the number of pandas four 
by integrating information across multiple frames, unlike the baseline, which failed. In Figure \ref{fig:qual}(c), when generating captions, our model provided more fine-grained details, capturing information that the baseline overlooked, even though it filtered spatial details.

Both the comprehensive quantitative and qualitative evaluations show that our approach delivers strong performance across various tasks while significantly improving efficiency, making it a competitive solution for untrimmed holistic video understanding.

\subsection{Model Analysis}

In this section, we conduct ablation studies on the Breakfast dataset for untrimmed video classification to evaluate the effectiveness of different components within the proposed framework. These studies provide insights into the contribution of each component to the overall model performance.

\noindent \textbf{Impact of Proposed Components.}
Table \ref{Tab:comp} demonstrates the impact of our newly introduced components, Relevance-Aware Temporal Compression (RTC) and Spatial Token Filtering (STF), on performance. Replacing the baseline MBC \cite{he2024ma} with RTC results in a significant accuracy improvement, with only a slight increase in computational overhead, which is almost negligible due to the lightweight scorer network in RTC. When STF is applied alongside the baseline MBC method, accuracy increases substantially, and the computational overhead is significantly reduced by filtering out redundant spatial tokens. Combining RTC and STF delivers the best performance while maintaining a notably low computational overhead.

\begin{table}

\caption{Ablation study results on the proposed components using the Breakfast dataset for untrimmed video classification.}
\vspace{-0.2cm}
\label{Tab:comp}
\centering
\resizebox{0.4\textwidth}{!}{
\begin{tabular}{l | c | l  } 
 \toprule
  Model &  Acc. (\%) $\uparrow$ & GFLOPs $\downarrow$ ($\Delta$) \\ [0.5ex] 
 \hline
    MBC (Baseline) & 91.6 & 154.7 (\textcolor{blue}{-}) \\
    RTC & 92.2 & 155.2 (\textcolor{blue}{0.004\% $\uparrow$}) \\
    MBC + STF & 92.9 & \textbf{101.7} (\textcolor{blue}{34.3\% $\downarrow$}) \\
    RTC + STF (Proposed) & \textbf{93.8} & 102.3 (\textcolor{blue}{33.9\% $\downarrow$})\\

\bottomrule
\end{tabular}
}
\vspace{-0.2cm}
\end{table}

\begin{table}

\caption{Comparison of different temporal modeling methods for untrimmed video classification using the Breakfast dataset.}
\vspace{-0.2cm}
\label{Tab:temp}
\centering
\resizebox{0.25\textwidth}{!}{
\begin{tabular}{l | c | l  } 
 \toprule
  Method &  Acc. (\%) $\uparrow$ & GFLOPs $\downarrow$ \\ [0.5ex] 
 \hline
    FIFO & 87.6 & 101.7 \\
    Avg. Pool. & 86.5 & 101.7  \\
    Linear Proj. & 87.0 & 102.1  \\
    Trans. Dec. & 93.5 & 308.3  \\
    TTS & 85.9 & 102.3 \\
    MBC & 92.9 & 101.7 \\
    \textbf{RTC (Ours)} & \textbf{93.8} & 102.3 \\

\bottomrule
\end{tabular}
}
\vspace{-0.6cm}
\end{table}

\noindent \textbf{Comparison of Temporal Modeling Methods}
Table \ref{Tab:temp} presents a comparison between our proposed relevance-aware temporal compression method and other temporal modeling approaches. In this experiment, we replaced the temporal modeling component within our framework and found that our method achieved the highest accuracy. The GFLOPs across all methods remainednearly identical and significantly low due to the use of STF, with the exception of the transformer decoder with learnable tokens \cite{xu2021long,reza2022history}. This approach skips the initial temporal compression step and directly processes the frame projections, resulting in a significantly higher computational cost despite still employing STF.

\begin{figure}[htbp]
	\centerline{\includegraphics[width=0.8\linewidth]{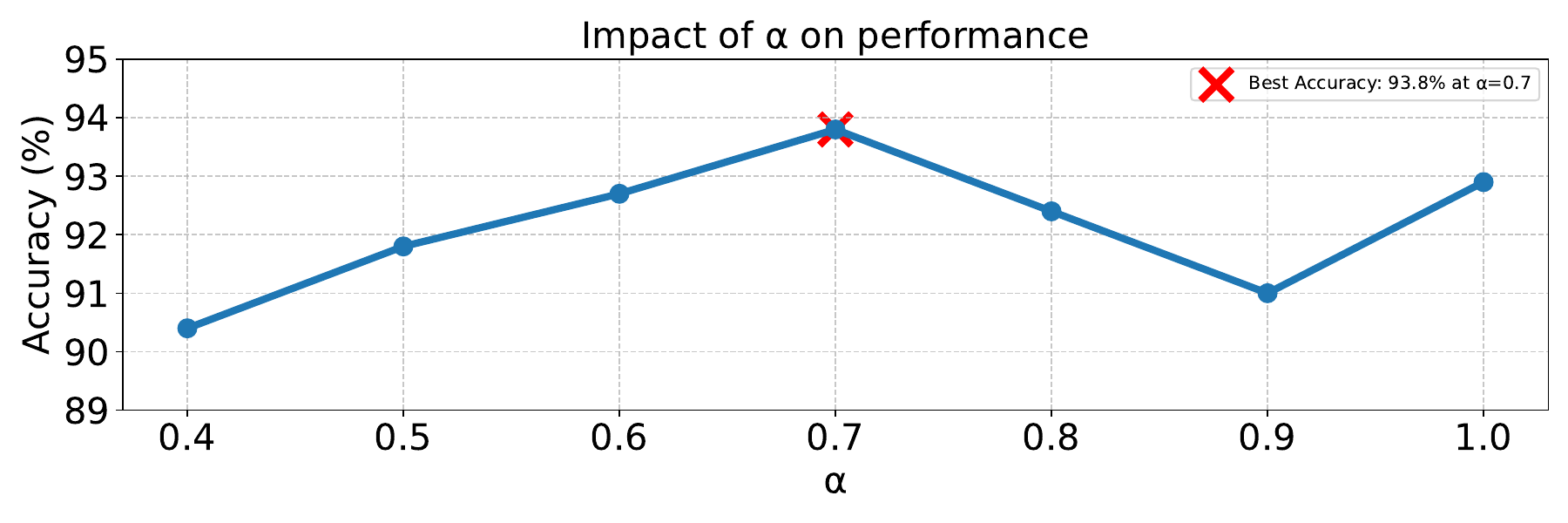}}
    \vspace{-4mm}
    \caption{Impact of score balance weight ($\alpha$) on untrimmed video classification performance in the Breakfast dataset.} \label{fig:alpha}
    \vspace{-1mm}
\end{figure}

\begin{figure}[htbp]
	\centerline{\includegraphics[width=0.8\linewidth]{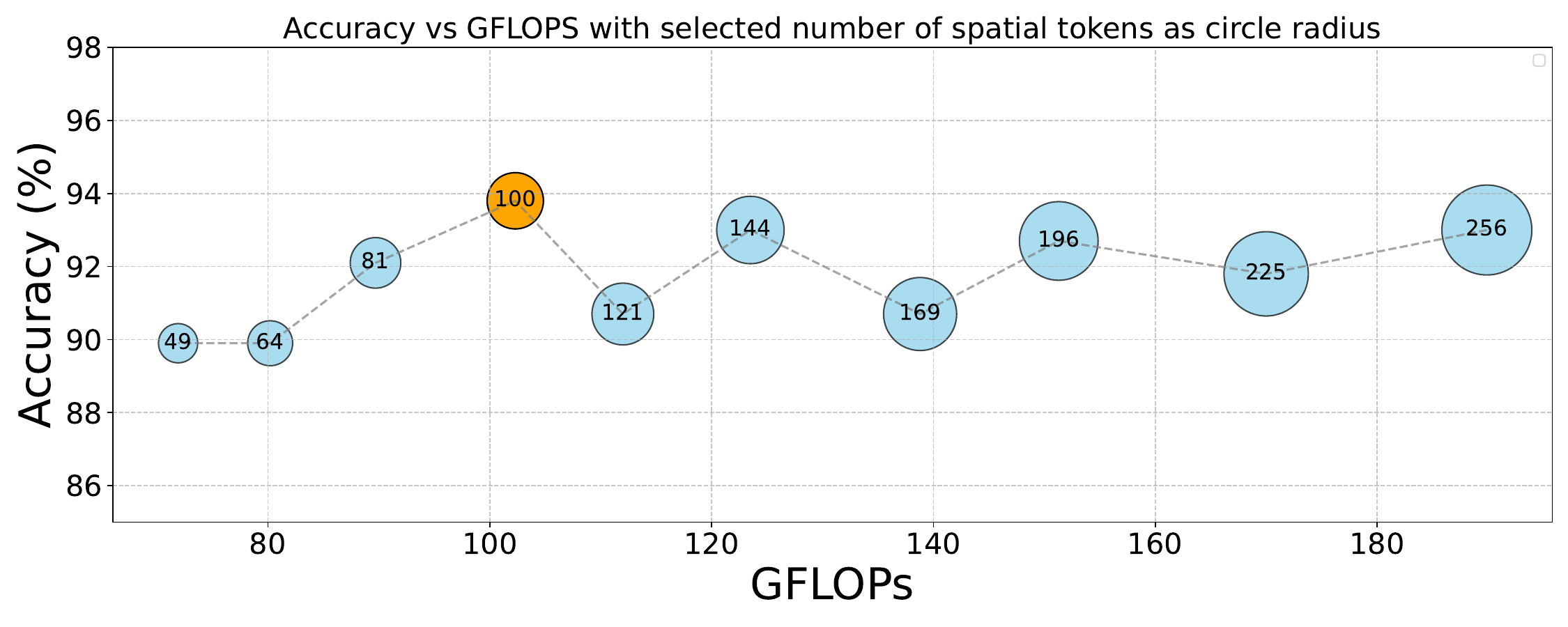}}
    \vspace{-4mm}
    \caption{Impact of selected spatial tokens on untrimmed video classification accuracy and GFLOP efficiency in the Breakfast dataset, highlighting the trade-off between quality and efficiency. } \label{fig:gflop}
    \vspace{-1mm}
\end{figure}

\noindent \textbf{Impact of Different Hyperparameters.}
In our framework, we tune various hyperparameters, focusing specifically on those related to the newly introduced component in this experiment. In the RTC module, we use the score balance coefficient \(\alpha\) to balance between the similarity-based score and the relevance-based score for selecting token pairs to merge (see Equation \ref{eqn:alpha}). As shown in Figure \ref{fig:alpha}, reducing \(\alpha\) from 1 does not follow a consistent pattern in accuracy improvement. However, the best accuracy is achieved at \(\alpha = 0.7\), after which the performance declines significantly, particularly at \(\alpha = 0.4\), leading us to stop further reduction. In Figure \ref{fig:gflop}, we examine the effect of the number of selected spatial tokens in the STF module. While accuracy does not follow a strict pattern, the model achieves optimal performance with 100 tokens before a significant drop occurs.




\section{Conclusion}


We introduced REEF, an efficient and relevance-aware LLM adapter designed for video-level understanding of untrimmed videos. Our approach incorporates two key components: (1) a context-aware temporal compression module that leverages relevance scores for improved visual memory compression, and (2) a spatial token filtering mechanism that eliminates redundant spatial tokens, significantly enhancing computational efficiency. REEF achieves state-of-the-art performance across four large-scale video datasets and three key benchmark tasks while maintaining low computational overhead.


Despite the promising results, several directions remain for future work. The token scoring and selection strategy could benefit from prompt-guided refinement. REEF’s online processing capability also makes it suitable for frame-by-frame online video understanding \cite{xu2021long, reza2022history}, though efficiency improvements are needed. Moreover, extending the framework beyond video-level tasks to support instance-level, frame-level, and pixel-level video understanding \cite{reza2022history, reza2023enhancing, ding2023mose, ding2023mevis, ying2023ctvis} would broaden its applicability across diverse video tasks.


\pagebreak
{
    \small
    \bibliographystyle{ieeenat_fullname}
    \bibliography{main}
}

\input{sec/X_suppl}

\end{document}

%% file: sec/X_suppl.tex
\clearpage
\maketitlesupplementary

\section{Differential Top-K Operation}
\label{app:topk}
In the main paper, we explain how the perturbed maximum method \cite{berthet2020learning, wang2022efficient, cordonnier2021differentiable} is used to make the Top-K operator differentiable, enabling the training of the token scorer networks. Building on this, we solve Equation \ref{eqn:perturbed}, and as outlined in \cite{wang2022efficient}, we carry out the following forward and backward operations to enable end-to-end training of the scorer networks with Top-K operations.

\paragraph{Forward Pass} A smooth approximation of the Top-K operation from Equation \ref{eqn:perturbed} can be implemented by introducing random perturbations and taking the expectation over these perturbations:

\begin{align}
\mathbf{Y}_{\sigma} = \mathbb{E}_{\mathbf{Z}} \left[ \arg \max_{\mathbf{Y} \in \mathcal{C}} \left\langle \mathbf{Y}, \mathbf{r'} \mathbf{1}^{\top} + \sigma \mathbf{Z} \right\rangle \right]
\end{align}

where \( \mathbf{Z} \) denote a random noise vector drawn from a standard Gaussian distribution, with \( \sigma \) as the hyperparameter controlling the noise variance. The symbol \( \langle \cdot, \cdot \rangle \) represents the inner product operation throughout the main paper and supplementary material. In practice, we execute the Top-K operation \( n \) times (where \( n = 500 \), as determined empirically to perform effectively across all experiments) and compute the average across these iterations.

\paragraph{Backward Pass} As described in \cite{cordonnier2021differentiable, wang2022efficient}, the Jacobian for the forward pass can be computed as:

\begin{align}
J_s\mathbf{Y} = \mathbb{E}_{\mathbf{Z}} \left[ \arg \max_{\mathbf{Y} \in \mathcal{C}} \left\langle \mathbf{Y}, \mathbf{r'} \mathbf{1}^{\top} + \sigma \mathbf{Z} \right\rangle \mathbf{Z}^\top / \sigma \right]
\end{align}

This expression simplifies in the case where \( \mathbf{Z} \) is normally distributed, allowing for efficient backpropagation through the Top-K operation.

We train the backbone models along with the token selection networks using a cross-entropy loss (described in equation \textcolor{red}{2} in the main paper). During training, the learned Top-K operation (implemented in PyTorch \cite{paszke2019pytorch}) is applied in the forward pass to enhance performance. By continuously applying random noise, the network becomes aware of the Top-K operation, allowing it to improve over time.

\section{Query Memory Bank}
\label{app:qmb}
Query Memory Bank (QMB) differs from the fixed visual memory bank, which stores static visual features. Instead, it accumulates input queries over time, denoted as \( \boldsymbol{\Theta}_t = \text{Concat}[\boldsymbol{\theta}_1, \boldsymbol{\theta}_2, \dots, \boldsymbol{\theta}_t], \boldsymbol{\Theta}_t \in \mathbb{R}^{tL \times D} \), creating a dynamic memory that captures the model’s evolving understanding of each frame up to the current timestep through the Q-Former. This Query Memory Bank also serves as the key and value: \( \mathbf{Q} = \boldsymbol{\theta}_t \mathbf{W}_Q, \quad \mathbf{K} = \boldsymbol{\Theta}_t \mathbf{W}_K, \quad \mathbf{V} = \boldsymbol{\Theta}_t \mathbf{W}_V \) as previously described. At each timestep, the learned query \( \boldsymbol{\theta}_t \) encapsulates critical information specific to the video up to that point. Unlike the static visual memory bank, these input queries \( \boldsymbol{\theta}_t \) are progressively refined through cascaded Q-Former blocks during training, enabling the capture of distinct video concepts and patterns at increasingly abstract levels. As a result, each self-attention layer is associated with a unique Query Memory Bank, where the stored queries are continuously updated during training.

\section{GFLOPs Calculation}
\label{app:gflops}
To evaluate our method's computational efficiency, we used giga floating-point operations (GFLOPs), a common metric for assessing efficiency in machine learning models. We employed the \textit{ptflop} python package \cite{ptflops} to calculate the GFLOPs for different components of our method. As noted in the main paper, the GFLOPs presented in the results tables are specific to the LLM adapter for a single pass. They exclude the LLM itself and are estimates, as certain components were manually calculated. Since we used the same LLM model with the same input token size as the baseline and focused exclusively on improving the adapter, we reported only the measurements related to the adapter for a fair comparison and to enhance the reader's convenience.

It is important to note that the estimated GFLOPs measurement for Video-Llama \cite{zhang2023video} is not directly comparable to MA-LMM or our approach, as Video-Llama processes the entire video in one pass without requiring an online iterative process. However, we included it for reference. Additionally, \textbf{`N/A'} is indicated for non-LLM methods in the results tables since they do not involve an LLM adapter, and we only reported GFLOPs for the LLM adapter.

\section{Experiment Configurations} 
\label{app:config}
The hyperparameters used in our experiments across different tasks and datasets are outlined in Tables \ref{Tab:1}, \ref{Tab:2}, and \ref{Tab:3}, detailing the configurations for untrimmed video classification, video question answering, and video captioning, respectively. These hyperparameters were determined empirically through systematic experimentation. For further details, the configuration files are available on GitHub alongside the codebase.

\begin{table}[!b]

\caption{Hyperparameters used in experiments across different datasets for the untrimmed video classification task.}
\label{Tab:1}
\centering
\begin{tabular}{l | P{0.2\linewidth}  P{0.2\linewidth}} 
 \toprule
  \textbf{Datasets} & \textbf{Breakfast} & \textbf{ActivityNet} \\ [0.5ex] 

 \hline
    LLM & \multicolumn{2}{c}{Vicuna-7B} \\
    Initial Training Epoch & \multicolumn{2}{c}{5} \\
    Main Training Epoch & \multicolumn{2}{c}{20} \\
    Learning Rate & \multicolumn{2}{c}{1e-4} \\
    Batch Size & \multicolumn{2}{c}{64} \\
    AdamW \(\beta\) & \multicolumn{2}{c}{(0.9, 0.999)} \\
    Weight Decay & \multicolumn{2}{c}{0.05} \\
    Image Resolution & \multicolumn{2}{c}{224} \\
    Beam Size & \multicolumn{2}{c}{5} \\
    Frame Length & \multicolumn{2}{c}{20} \\
    Memory Bank Length & \multicolumn{2}{c}{10} \\
    Score Balance Weight \(\alpha\) & 0.7 & 0.9 \\
    Selected Spatial Tokens & 100 & 144 \\
    Prompt & “What type of breakfast is shown in the video?” & “what is the person doing in the video?” \\
 
\bottomrule
\end{tabular}
\end{table}

\begin{table}

\caption{Hyperparameters used in experiments across different datasets for the video question answering task.}
\label{Tab:2}
\centering
\begin{tabular}{l | P{0.2\linewidth}  P{0.2\linewidth}} 
 \toprule
  \textbf{Datasets} & \textbf{MSVD} & \textbf{ActivityNet} \\ [0.5ex] 

 \hline
    LLM & \multicolumn{2}{c}{Vicuna-7B} \\
    Initial Training Epoch & \multicolumn{2}{c}{2} \\
    Main Training Epoch & \multicolumn{2}{c}{5} \\
    Learning Rate & \multicolumn{2}{c}{1e-4} \\
    Batch Size & \multicolumn{2}{c}{128} \\
    AdamW \(\beta\) & \multicolumn{2}{c}{(0.9, 0.999)} \\
    Weight Decay & \multicolumn{2}{c}{0.05} \\
    Image Resolution & \multicolumn{2}{c}{224} \\
    Beam Size & \multicolumn{2}{c}{5} \\
    Frame Length & \multicolumn{2}{c}{20} \\
    Memory Bank Length & \multicolumn{2}{c}{10} \\
    Score Balance Weight \(\alpha\) & 0.7 & 0.9 \\
    Selected Spatial Tokens & 100 & 121 \\
    Prompt & \multicolumn{2}{c}{“Question: \{\} Short Answer:”} \\
 
\bottomrule
\end{tabular}
\end{table}

\begin{table}

\caption{Hyperparameters used in experiments across different datasets for the video captioning task.}
\label{Tab:3}
\centering
\begin{tabular}{l | P{0.2\linewidth}  P{0.2\linewidth}} 
 \toprule
  \textbf{Datasets} & \textbf{MSVD} & \textbf{YouCook2} \\ [0.5ex] 

 \hline
    LLM & \multicolumn{2}{c}{Vicuna-7B} \\
    Initial Training Epoch & \multicolumn{2}{c}{3} \\
    Main Training Epoch & \multicolumn{2}{c}{10} \\
    Learning Rate & 1e-5 & 1e-4 \\
    Batch Size & \multicolumn{2}{c}{96} \\
    AdamW \(\beta\) & \multicolumn{2}{c}{(0.9, 0.999)} \\
    Weight Decay & \multicolumn{2}{c}{0.05} \\
    Image Resolution & \multicolumn{2}{c}{224} \\
    Beam Size & \multicolumn{2}{c}{5} \\
    Frame Length & \multicolumn{2}{c}{80} \\
    Memory Bank Length & \multicolumn{2}{c}{40} \\
    Score Balance Weight \(\alpha\) & \multicolumn{2}{c}{0.7} \\
    Selected Spatial Tokens & 100 & 144 \\
    Prompt & \multicolumn{2}{c}{“what does the video describe?”} \\
 
\bottomrule
\end{tabular}
\end{table}